\documentclass[journal]{IEEEtran}
\usepackage[left=2cm,right=2cm,top=2cm,bottom=3cm]{geometry}
\usepackage{amsmath}
\usepackage{rotating}
\usepackage{color}
\usepackage{longtable}
\usepackage[normalem]{ulem}

\begin{document}
{\huge \textbf{IEEE copy right notice}}

©2020 IEEE.  Personal use of this material is permitted.  Permission from IEEE must be obtained for all other uses, in any current or future media, including reprinting/republishing this material for advertising or promotional purposes, creating new collective works, for resale or redistribution to servers or lists, or reuse of any copyrighted component of this work in other works.

Accepted to be published in: IEEE Transactions on Biomedical Engineering

\newpage

\title{Cable-driven robotic interface\\for lower limb neuromechanics identification}

\author{Hsien-Yung Huang, Ildar Farkhatdinov, Arash Arami, Mohamed Bouri and Etienne Burdet
\thanks{Email: \{h.huang14, e.burdet\}@imperial.ac.uk. All authors but Bouri are or were with the Department of Bioengineering, Imperial College of Science, Technology and Medicine, UK (https://www.imperial.ac.uk/human-robotics). Farkhatdinov is also with the School of Electronic Engineering and Computer Science, Queen Mary University of London, UK. Arami is with the Department of Mechanical and Mechatronics Engineering, University of Waterloo, Canada and with Toronto Rehabilitation Institute-University Health Network, Toronto, Canada. Bouri is with the Biorobotics Laboratory, \'Ecole Polytechnique F\'ed\'erale de Lausanne. We thank Jonathan Eden for editing the text. This work was funded in part by the EC grants: ICT-601003 BALANCE, ICT-611626 SYMBITRON, and H2020 ICT-871767 REHYB.}
}
\maketitle

%%%%%%%%%%%%%%%%%%%%%%%%%%%%%%%%%%%%%%%%%%%%
\begin{abstract}
This paper presents a versatile cable-driven robotic interface to investigate the single-joint joint neuromechanics of the hip, knee and ankle in the sagittal plane. This endpoint-based interface offers highly dynamic interaction and accurate position control (as is typically required for neuromechanics identification), and provides measurements of position, interaction force and electromyography (EMG) of leg muscles. It can be used with the subject upright, corresponding to a natural posture during walking or standing, and does not impose kinematic constraints on a joint, in contrast to existing interfaces. Mechanical evaluations demonstrated that the interface yields a rigidity above 500\,N/m with low viscosity. Tests with a rigid dummy leg and linear springs show that it can identify the mechanical impedance of a limb accurately. A smooth perturbation is developed and tested with a human subject, which can be used to estimate the hip neuromechanics.
\end{abstract}
%%%%%%%%%%%%%%%%%%%%%%%%%%%%%%%%%%%%%%%%%%%%
\section{Introduction}
\label{ch:Introduction}
An accurate characterisation of lower limb neuromechanics is required to understand lower limb neurophysiology and to design appropriate control for robotic walking aids. To identify the limbs neuromechanics, a rigid robotic interface equipped with powerful actuators is typically required to apply force/position disturbances while sensors record the resulting modification of position/force. Available robotic interfaces to identify the lower limb neuromechanics include motor-driven dynamometers \cite{Sinclair2006, Hahn2011, Amankwah2004, Dirnberger2012b, Alvares2015, Valovich-mcLeod2004} and gait rehabilitation exoskeleton devices \cite{Lunenburger2005, Veneman2007}, whose characteristics are compared in Table \ref{tab:TableOfDevice}.

\subsection{Existing neuromechanics estimation devices}
\label{Existing neuromechanics estimation devices}
Motor-driven dynamometers are robotic interfaces that perform single joint rotations while the two limbs are fixed to the interface. Such interfaces can be used for single joint identification and physical therapy, providing isotonic, isometric and isokinetic experimental conditions. They have been used to estimate the torque-angle relation of the ankle joint \cite{Hahn2011}, injury- or disease-induced increase in ankle joint stiffness \cite{Lorentzen2010}, passive resistance torque increased at the knee \cite{Perell1996} or at all joints \cite{Akman1999} due to spinal cord injury (i.e. spasticity), and the difference in ankle range of motion (ROM) in individuals with cerebral palsy \cite{DeGooijer-VanDeGroep2013, Sloot2015}. These interfaces, however, impose movement constraints on a joint, which may result in unnatural motions, and accounts for variability in torque-angle identification results using single-joint motor-driven dynamometer relative to estimations from multi-joint inverse dynamics \cite{Hahn2011}. Furthermore, motor-driven dynamometers are not equipped with body weight support for hip joint measurements. Therefore, hip joint neuromechanics investigations with those interfaces could only involve healthy participants \cite{Claiborne2009}, or be used with impaired individuals in postures not requiring weight bearing. 

Gait rehabilitation robotic exoskeletons are interfaces affixed to the body. They can provide controlled gait assistance \cite{Lunenburger2005, Veneman2007, Jin2015, Vouga2017} and be used to analyse neuromechanical factors such as spasticity and voluntary muscle force level \cite{Lunenburger2005}, or joint impedance during leg swinging \cite{Koopman2016}. However, exoskeletons constrain the joints movement, and may induce non-negligible vibrations due to the difficulty to design a rigid mechanical structure. Furthermore, they may not be suitable to develop specific experiment protocols required to measure the lower limb neuromechanics. For example, the stiffness estimation method of \cite{Mirbagheri2000, Mirbagheri2001} relies on perturbation pulse trains of 1.72$^\circ$ amplitude and 150\,ms pulse width, which is challenging to implement on existing gait rehabilitation exoskeletons due to their limited rigidity and torque capabilities.

Most existing devices focused on ankle \cite{Rouse2014,Chung2004,Shorter2019} or knee joint measurements, while devices targeting the hip joint could usually be used only for isokinetic motion \cite{Claiborne2009, Bieryla2009} or for multi-joint torque perturbations \cite{Koopman2016} that may limit the accuracy of the measurements. In view of these functional limitations, this paper presents a robotic interface that can be used to systematically investigate the single-joint neuromechanics of the hip, knee and ankle joint in a natural upright position, without constraining the targeted joint motion, and with negligible structural vibrations even during highly dynamic perturbations. This interface is validated on hip joint viscoelasticity estimation in one subject.

%table
\begin{table*}[]
\caption{Characteristics of existing lower limb neuromechanics evaluation devices}
\label{tab:TableOfDevice}
\centering
\begin{tabular}{|p{0.75\columnwidth}|p{0.24\columnwidth}|p{0.21\columnwidth}|p{0.34\columnwidth}|p{0.16\columnwidth}|}
\hline
{\bf device name} & {\bf max force/torque} & {\bf speed} & {\bf characteristics} & {\bf relevant papers}\\ \hline

IsoMed2000 (D \& R Ferstl GmbH, Germany)& 500\,N\,/\,750\,Nm & 1\,-\,560\,$^{\circ}$\!/s & Single-joint direct drive (hip, knee and ankle) & \cite{Hahn2011, Dirnberger2012a}
%\begin{tabular}[c]{@{}l@{}} 2,4,8
%\cite{Hahn2011,Dirnberger2012b,Dirnberger2012\end{tabular} 
\\ \hline

%no name (MOOG FCS Inc., Netherlands)& unknown & unknown & Ankle joint direct drive & \cite{DeGooijer-VanDeGroep2013, Sloot2015} %\begin{tabular}[c]{@{}l@{}}\cite{DeGooijer-VanDeGroep2013, Sloot2015} \end{tabular} 
%\\ \hline

Biodex 3 dynamometer (Biodex Medical Systems, USA)& 890\,N & 0\,-\,450\,$^{\circ}$\!/s & Single-joint direct drive (hip, knee and ankle) & \cite{Claiborne2009, Bieryla2009} %\begin{tabular}[c]{@{}l@{}}\cite{Sinclair2006, Amankwah2004, Alvares2015, Valovich-mcLeod2004, Claiborne2009, Bieryla2009}\end{tabular} 
\\ \hline

Cybex 770 Norm (Lumex Inc., USA)& 678\,Nm & 0\,-\,500\,$^{\circ}$\!/s & Single-joint direct drive  (hip, knee and ankle) & \cite{Alvares2015,Perell1996} 
%\cite{Alvares2015,Perell1996} 
\\ \hline

Kin-Com model 500H (Charracx Co., TN, USA)& unknown & unknown & Single-joint direct drive & \cite{Akman1999}  %\cite{Akman1999} 
\\ \hline

Lokomat (Hocoma AG, Switzerland)& unknown & 0.89\,m/s & Gait rehabilitation device & \cite{Lunenburger2005} %\begin{tabular}[c]{@{}l@{}}\cite{Lunenburger2005}\end{tabular} 
 \\ \hline

LOPES I (University of Twente \& Moog, Netherlands)& 250\,N\,/\,50\,Nm & allows walking up to 5\,km/h & Gait research device & \cite{Veneman2007, Koopman2016}
\\ \hline

LOPES II (University of Twente \& Moog, Netherlands)& 250\,N\,/\,60\,Nm & allows walking up to 5\,km/h & Gait research device & \cite{Meuleman2016}
%\begin{tabular}[c]{@{}l@{}}\cite{Koopman2016, Meuleman2016}\end{tabular} 
\\ \hline

\end{tabular}
\end{table*}

\subsection{Functional requirements}
\label{ch:Functional requirements}
We want to develop a versatile device with a powerful actuator, a rigid mechanical structure and suitable control, that can be used to implement various protocols, including isometric and isokinetic conditions as well as brief mechanical perturbations, in order to characterise the lower body joints neuromechanics. To implement isometric conditions, the device should be able to provide a strong standstill torque to resist the human subject's maximum voluntary contractions. The joint torque measurements obtained during a stair climbing experiment \cite{Andriacchi1980} were used as a reference for our device's actuator's standstill torque. To perform isokinetic experiments and quantify velocity-dependent muscle or joint behaviour, our device should also be able to implement a full range of movements with constant velocity range from 0 to 250$^\circ$/s \cite{Fornusek2007}. Finally, the device should be able to realise the highly dynamic movements required for estimating mechanical joint impedance. For instance, evaluating ankle reflexes as in \cite{Mirbagheri2000, Mirbagheri2001} requires a fast ankle angle perturbation with 1.72$^\circ$ amplitude and 150\,ms pulse width. In order to perform similar controlled position perturbations on the hip joint, considering that a human leg contributes to 15-20\% of the body weight \cite{Winter2009}, a torque up to 100\,Nm would be required for a 90kg subject. 

Large forces and accelerations required to characterise the lower limb neuromechanics demand a powerful and thus heavy actuator. Therefore, this actuator should be rigidly fixed away from the moving limb and reliable motion transmission should be used. A pneumatic cylinder connection \cite{Aoyagi2007} or a two-bar linkage device \cite{Fujii2007} could be used in this purpose, which would however result in a limited workspace. A pretensioned cable transmission can provide actuation with low inertia and without backlash \cite{Townsend1988}. 

The structure of the interface should be rigid, to minimise undesired vibrations and deformations. An additional structure is required to support the subject in an upright posture such as for walking. This structure should be mechanically independent from the robotic interface in order to avoid the transmission of vibrations from the actuator. Finally, to avoid constraining the hip and knee joints motion such as with the Lokomat \cite{Lunenburger2005} and LOPES \cite{Koopman2016} exoskeletons, we developed an end-point based interface interacting with the extremity of the examined limb while the rest of the body can move freely. This paper presents and validates this {\it Neuromechanics Evaluation Device} (NED). Section~\ref{ch:Device design} explains its design concept, its components and control. Section~\ref{ch:System characterisation} analyses the resulting kinematics and parameters sensitivity, with mechanical characterisations. Typical applications are illustrated in section~\ref{ch:Validating}.

%%%%%%%%%%%%%%%%%%%%%%%%%%
\section{Device design}
\label{ch:Device design}

%figure
\begin{figure}[]
\centering
\includegraphics[width=0.90\columnwidth]{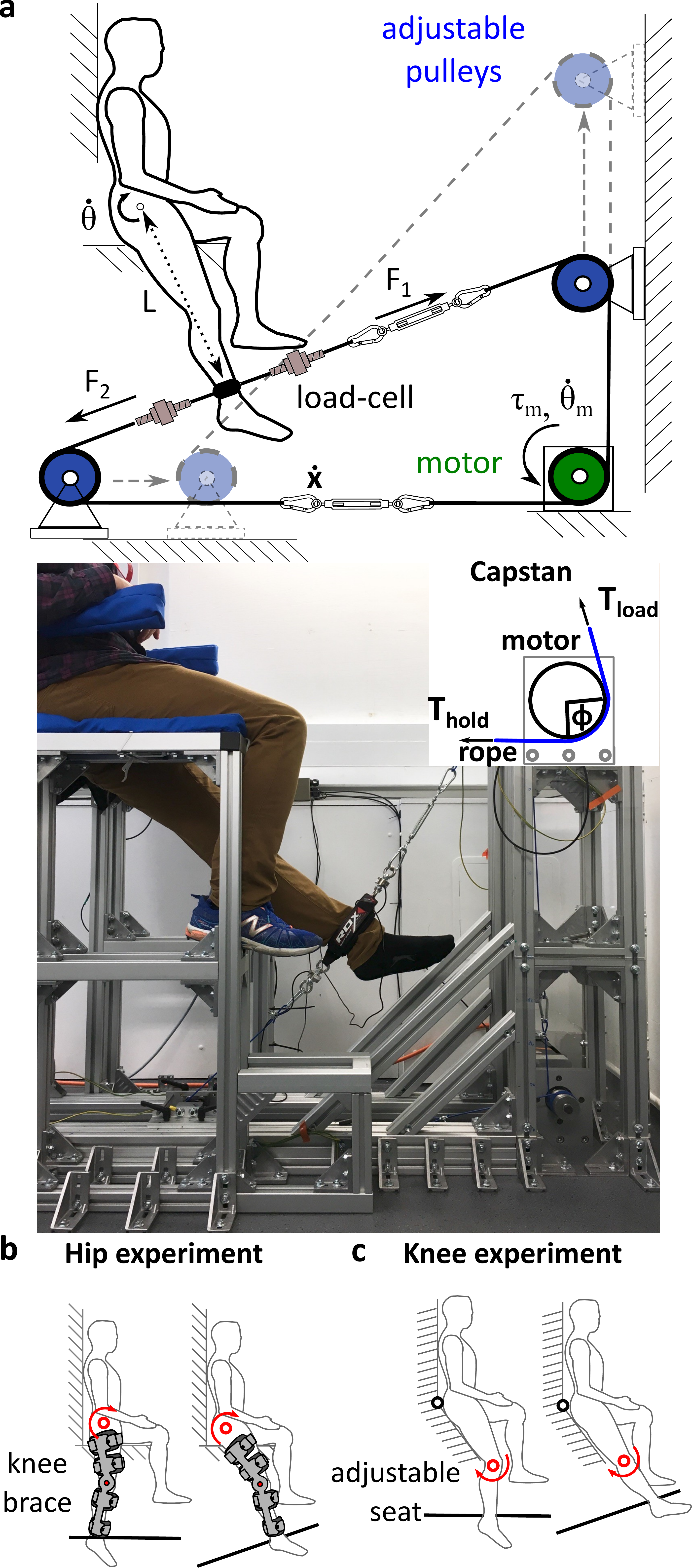}
\caption[Description of the Neuromechanics Evaluation Device (NED)]{\small Neuromechanics Evaluation Device (NED). Panel (a) shows how the subject seated in a rigid chair with an open design allowing the leg movement. The motor force transmitted by the cable is measured by load-cells on both sides of the ankle fixture (front force $F_1$, rear force $F_2$). $\dot{\theta}_m$ and $\tau_{m}$ are the speed and torque at the motor, $\dot{\theta}$ the hip joint angular velocity, $\dot{x}$ the cable linear movement speed and $L$ the measured leg length. The inset figure details the interaction force between pulley and the cable rope. The relation between the holding force $T_{hold}$ and the loading $T_{load}$ can be described as Equ.\ref{eq:capstan} with a contact angle of $\phi$ and a coefficient of friction $\mu$. Panel (b) and (c) show how NED can be used to carry out experiments to investigate the knee and hip neuromechanics. The front and back pulleys can be shifted along the rail to ensure a perpendicular cable interaction.}
\label{fig:NEDSketch}
\end{figure}

\subsection{General description}
\label{ch:General description}
Considering all the design factors presented in the previous section, our solution included a cable-driven device with a large actuator placed outside of the workspace transmitting power to the limb, while the subject is supported in a natural upright posture by an independent structure. The developed {\it Neuromechanics Evaluation Device} (NED) is illustrated in Fig.\,\ref{fig:NEDSketch}. 

As shown in Fig.\,\ref{fig:NEDSketch}a, the subject is half seated on a rigid chair (of length 0.55\,m, width 0.7\,m and height 1.5\,m) with one leg suspended in the workspace and attached to the system via a foot fixture. The leg is moved by the motor (AM8061, Beckhoff, Germany) located at the bottom of the workspace, via a steel cable (7x7 galvanised steel with PVC coating). Two load-cells (TAS510, HT sensors, China) are placed between the extremities of the foot fixture and the cable to measure the respective interaction forces. The front and rear pulleys can be locked at different positions along the rail (3\,m in the horizontal direction and 1.5\,m vertically) in order to keep the cable perpendicular to the leg (as shown in Fig.\,\ref{fig:NEDSketch}a) and tensed. The cable tension is adjusted by the turnbuckles placed in series with the cable. All structures are made with aluminum strut profiles (40x40L, Bosch Rexroth, Germany) and bolted to the cement floor.

The open seat enables the experimenter to perform hip experiments at different knee angles as shown in Fig.\,\ref{fig:NEDSketch}b. In this case, the knee joint can be kept at a specific joint angle using a knee brace (T-scope, Breg), which enables us to study the influence of the knee angle. By adjusting the seat, it becomes also possible to study the knee neuromechanics as shown in Fig.\,\ref{fig:NEDSketch}c.

In order to select an actuator based upon our design criteria, motors produced by Beckhoff, ETEL and Infranor were evaluated depending on their technical specifications including motor peak torque, standstill torque, moment of inertia and possible gearbox reduction ratio. The actuator selection process is detailed in \cite{Huang2019c}, along with NED's components list and their characteristics, including the selected actuator.
%%%%%%%%%%%%%%%%%%%%%%%%
\subsection{Cable transmission}
\label{ch:Cable transmission}
The cable transmission is designed to support a sufficiently large contact force between the cable and the pulley to prevent slippage at the desired actuation force. The Capstan equation 
\begin{equation}
\label{eq:capstan}
T_{load} = \, T_{hold} \, e^{\mu \phi}
\end{equation}
describes the force relation between the cable and the contact area of a cylinder while pulling a leg forward (inset Fig.\,\ref{fig:NEDSketch}a), where $T_{load}$ is the front cable tension which bares larger loading, $T_{hold}$ the force required to hold the loading on pulley, $\mu$ the friction coefficient between the cable and pulley, and $\phi$ the contact angle. The minimum contact angle $\phi$ to prevent slippage is calculated from the friction coefficient and the expected cable force on both sides of the pulley. Assuming a joint torque of 150\,Nm \cite{Andriacchi1980} and a plastic-metal friction coefficient of 0.1\,-\,0.3, there must be at least four cable turns. Conservatively, the cable was wounded five times around the pulley.
%%%%%%%%%%%%%%%%%%%%%%%%
\subsection{Control system}
\label{ch:Control system}
%figure
\begin{figure}[]
\centering
\includegraphics[width=\linewidth]{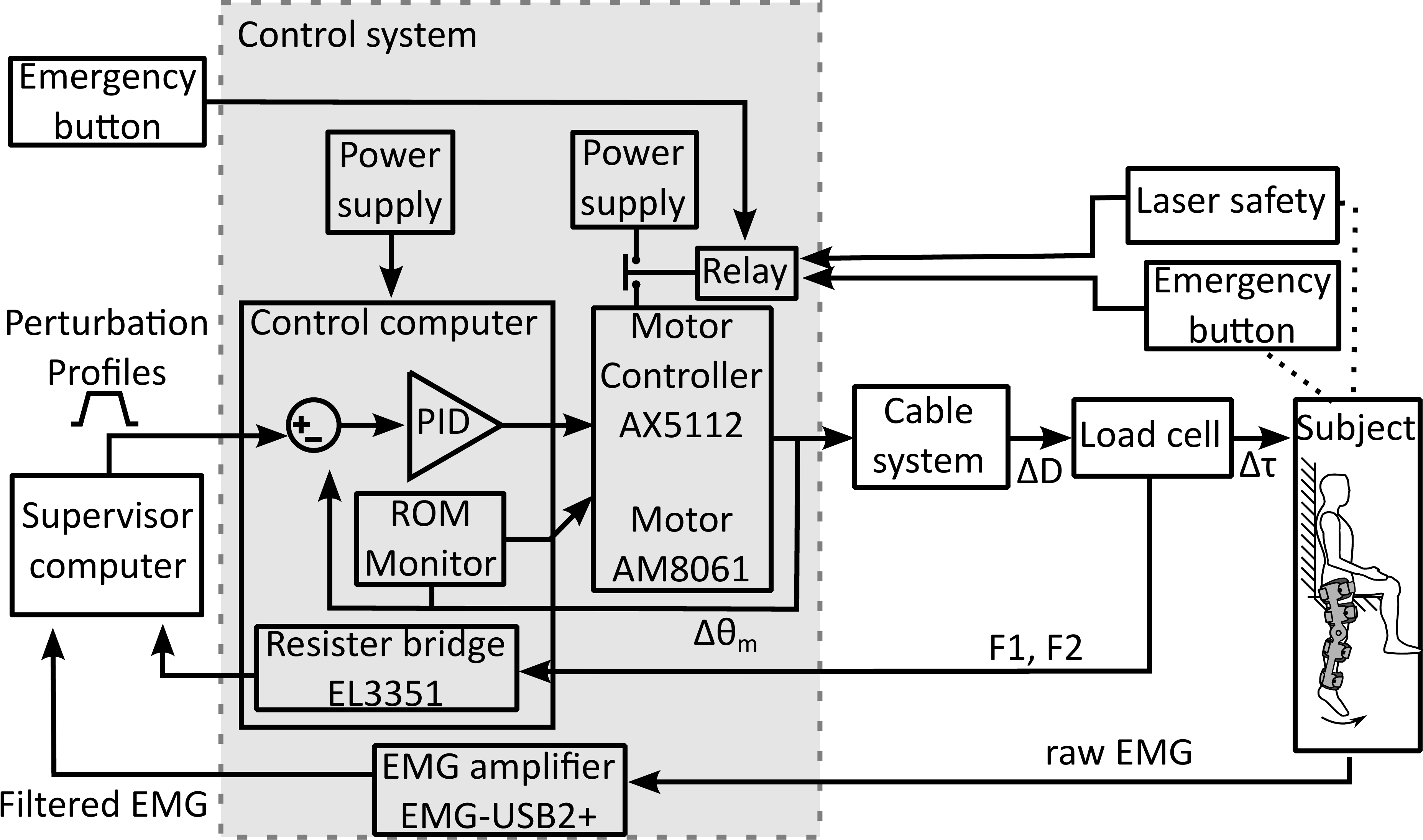}
\caption[The control system of NED]{\small The control system of NED. }
\label{fig:Architecture}
\end{figure}

The control architecture of NED is depicted in Fig.\,\ref{fig:Architecture}. The supervisor computer provides the position command to the control computer (CX5130, Beckhoff), which performs the real-time control and monitors the motor controller (AX5112, Beckhoff) and motor (AM8061, Beckhoff). The cable system transmits the motion. In the example of a position disturbance, the angular displacement of the motor shaft $\Delta \theta_m$ is monitored by the software limits to avoid over-stretching the leg. The two load-cells at the extremities of the ankle fixture record the interaction forces $F_1$, $F_2$, which are fed back via a resistor bridge unit. The activity of leg muscles are recorded with surface electromyography (EMG) electrodes and filtered. Safety relays are connected to the power supply of the motor controller; they are activated by a laser safety system and emergency buttons when a fault is detected. Both the control computer and motor controller are powered separately by a power supply (PRO ECO 120W, Weidmuller, Finland). The operating software environment TwinCAT plans and implements the fastest point-to-point motion with given speed, acceleration and jerk limits.

In addition to considering the feedback error, the motor controller uses three sensor channels. Two resistor bridges (EL3351, Beckhoff) are used to measure the interaction force between the device and the subject's leg, and an analog channel (EL3255, Beckhoff) can be used e.g. for measuring the leg motion. Adding a motion capture system would enable to measure hip joint rotations directly and could improve the leg displacement measurement relative to the current estimation from the motor encoder.

%figure
\begin{figure}[]
\centering
\includegraphics[width=\linewidth]{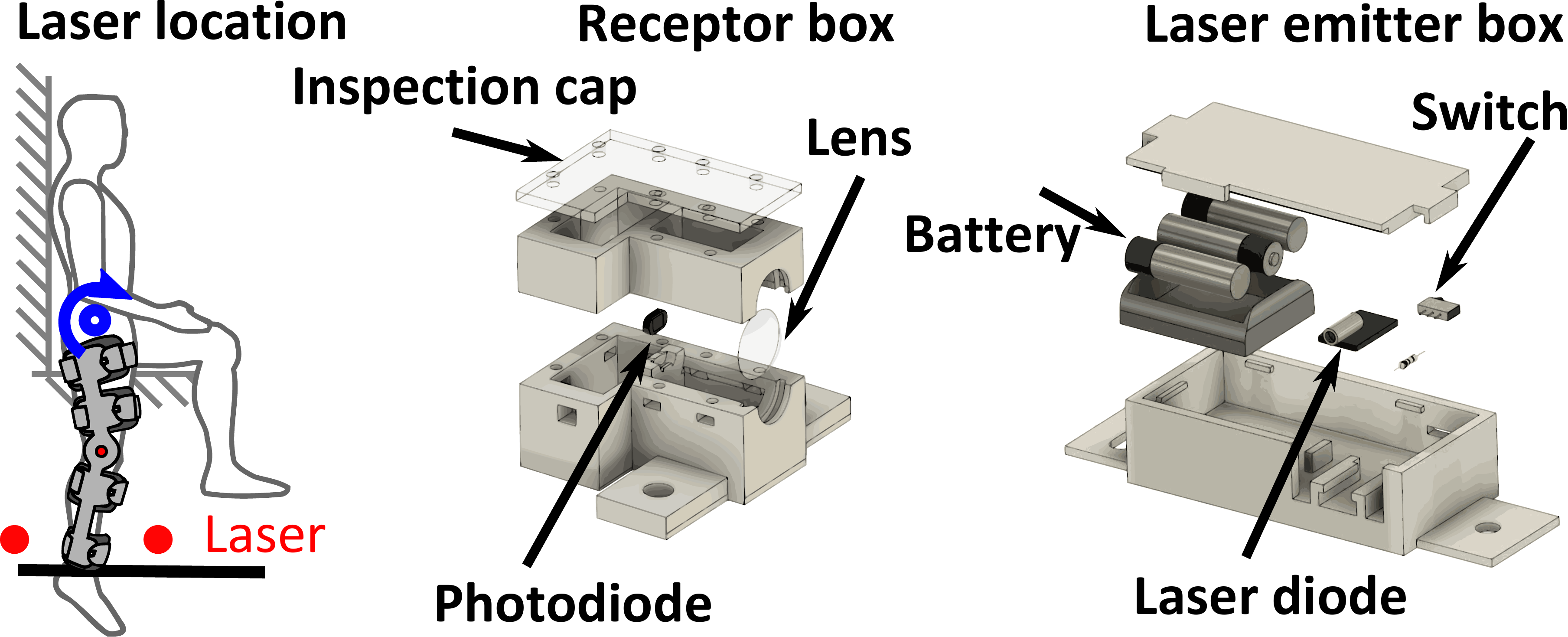}
\caption[Laser safety system]{\small Laser safety system composed of a laser emitter box (with a focusing lens) and a receptor box. Any obstacle blocking the laser transmission will immediately shut down the power supply to the motor controller.}
\label{fig:Laser}
\end{figure}
%%%%%%%%%%%%%%%%%%%%%%%%
\subsection{Safety measure and ergonomics}
\label{ch:Safety measure and ergonomics}
Safety is a critical factor for robotic interfaces which are in contact with the human body. Therefore, we implemented redundant hardware and software measures to ensure safety throughout our experiments. A safety system was developed to define the allowable range of motion as shown in Fig.\,\ref{fig:Laser}. The laser box emits a signal to the photodiode in the receptor box, which controls two safety relays. If the laser beam is blocked by any obstacle, e.g. leg moving beyond the expected range, safety relays will shut down the motor controller. Second, software safety measures implemented in the motor controller shut down the power when a position, speed, acceleration or power/torque limit is reached. Specific software limits define the workspace in which the leg should move depending on the targeted experiment. Finally, the power supply of both the motor and motor controller will shut down if any of the three emergency buttons is pressed. These buttons, which are available to the subject and experimenters during the experiment, are connected to another two safety relays and a master switch. In total, four safety relays (PSR-MS35, Phoenix Contact, Finland) control the power supply of the controller. If any safety threshold is reached the power motor is set off. 

Besides the safety measures described above, various factors are included to provide a comfortable environment for different subjects. The dimensions of NED, that includes rail lengths and chair size, are designed for human subjects of height between 1.5\,-\,1.8\,m. The rigid chair is covered with memory-foamed cushions to increase experiment contentment, and handrail location is adjustable to optimised body weight support. 

%%%%%%%%%%%%%%%%%%%%%%%%
\section{System characterisation}
\label{ch:System characterisation}

This section first analyses the kinematics of the developed system. It then examines sensor measurement errors and solutions to ensure an accurate recording. Lastly, a series of system identification tests are performed to characterise the system in different dynamical conditions.

%figure
\begin{figure}[]
\centering
\includegraphics[width= \columnwidth]{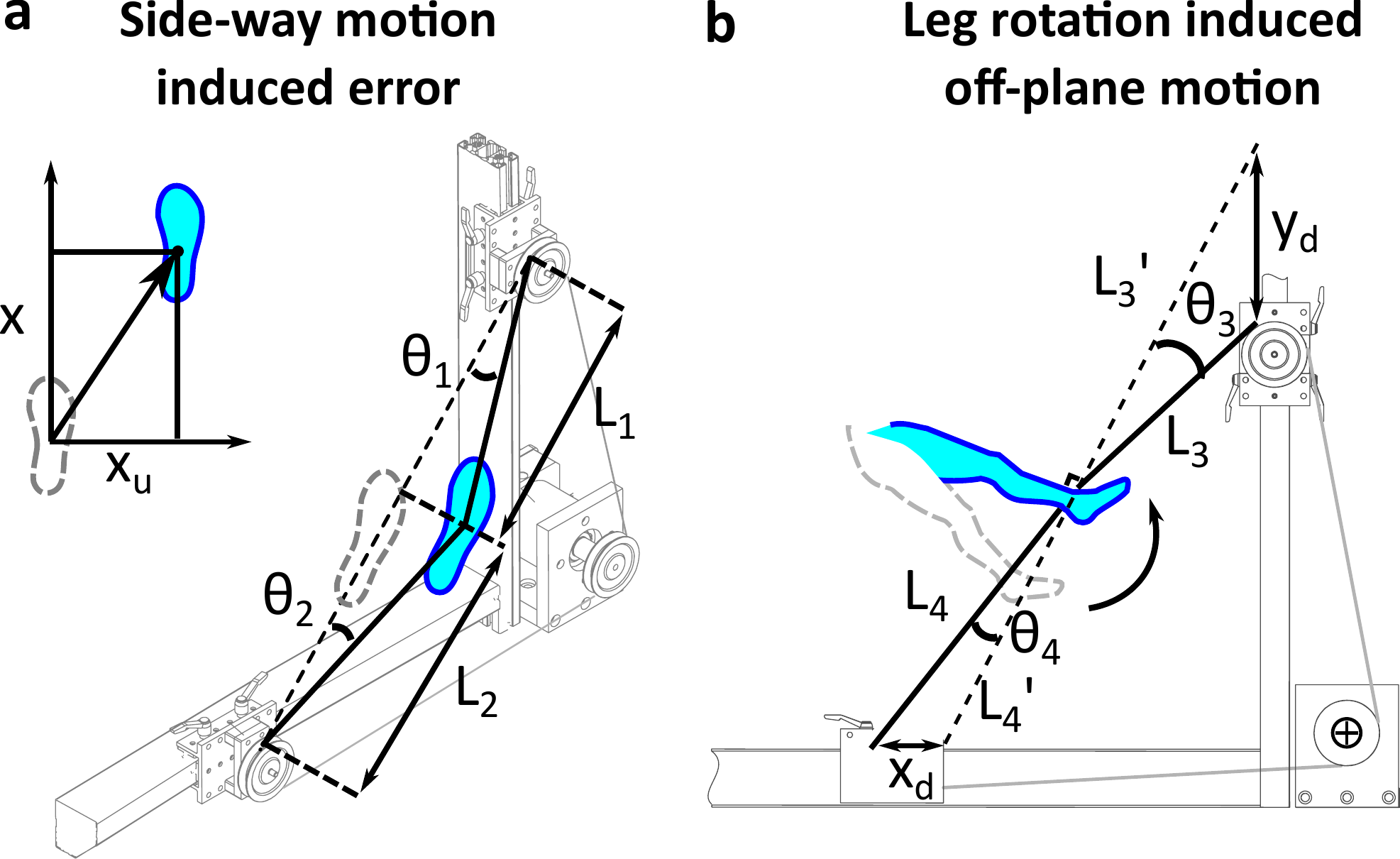}
\caption[Off-plane motion induced error]{\small Off-plane motion induced error. Panel (a) illustrates the measurement error resulting from side-way motions. $\theta_1$ and $\theta_2$ are the misalignments between the load-cell and the desired leg motion, $L_1$ and $L_2$ the distances between the foot and both pulleys. The maximum permissible side-way motion $x_u$ can be calculated by trigonometry. Panel (b) depicts the error resulting from a large leg rotation. Since both pulleys are fixed for each experiment, large leg motions will cause an angle between the load-cell and line of motion, which is described as angles $\theta_3$ and $\theta_4$. $L_3$ and $L_4$ are the distances between the foot and the pulleys. On the other hand, $L_3'$ and $L_4'$ are the distances between the foot and the ideal pulley location. $x_d$ and $y_d$ are the distances between the ideal pulley location and actual pulley location.}
\label{fig:SensitivityAnalysis}
\end{figure}

\subsection{Kinematics and the sensitivity analysis}
\label{ch:Kinematics and the sensitivity analysis}
NED is designed to measure the lower limb joints' flexion-extension biomechanics assuming that the cable and leg motions are restricted to the sagittal plane (with sufficient high cable pretension). For small angular displacements with the knee stretched and locked to maintain the leg straight, the kinematics is:
\begin{equation}
\label{eq:Jacobian_Simplified}
\rho_m \, \dot{\theta}_m = \dot{x} = L\, \dot{\theta} \quad
\to \quad \dot{\theta} = \frac{\rho_m}{L}\dot{\theta}_m \,
\end{equation}
where $\rho_m$ is the motor pulley diameter, $\dot{\theta}_m$ is the motor speed, $\dot{x}$ is the cable linear motion speed, $L$ the leg length and $\dot{\theta}$ the hip joint rotation speed.

To ensure that the aforementioned planar movement assumption is valid, we investigated how lateral leg movements can potentially influence the accuracy of the biomechanical measurements. We assume that during an experiment the leg-cable attachment point can displace sideways by an undesired distance $x_u$ measured from the normal plane of movement as shown in Fig.\,\ref{fig:SensitivityAnalysis}a. Then, the cable force measurements are affected by the off-plane configuration described by the angles $\theta_1$\,=\,arctan\,($x_u/L_1$) and $\theta_2$\,=\,arctan\,($x_u/L_2$) as shown in Fig.\,\ref{fig:SensitivityAnalysis}a. By considering different leg lengths [80\,-\,95]\,cm, hip angles [5$^\circ$-\,60$^\circ$] and different pulley locations ([80-165]\,cm horizontal and [45-110]\,cm vertical), it is shown that the side-way motion should be limited within 14.3\,cm to result in a force measurement error below 5\% (that may lead to a 5\% stiffness estimation error). A side-way motion test (with 200\,N cable pretension) shows that a 14\,cm side-way motion requires an external force of 225\,N. Experiments should therefore be limited within such force limitations.

As a large angular displacement cannot be considered as linear motion, we further investigated the influence of limb rotation upon measurement accuracy. For each experiment, the pulley locations are relocated and fixed to yield a perpendicular cable connection minimising the measurement errors (shown as the gray dashed line in Fig.\,\ref{fig:SensitivityAnalysis}b). Moving far away from the initial position will cause measurement error due to angles $\theta_3 \!=\! (L_3^2\!+\!L_3'^2\!-\!Y_d^2)/(2\,L_3\,L_3')$ and $\theta_4 \!=\! (L_4^2\!+\!L_4'^2\!-\!X_d^2)/(2\,L_4\,L_4')$ (the black lines). By considering different leg lengths [80\,-\,95]\,cm, different experiment hip angles [5\,$^{\circ}$\,-\,60\,$^{\circ}$] and different sizes of leg motions, the largest acceptable leg motion before reaching a 5\% error in both measurement and stiffness estimation is 21\,cm in either direction. This can be considered as maximum acceptable position displacement to design experiments.

%%%%%%%%%%%%%%%
%figure
\begin{figure}[]
\centering
\includegraphics[width=\linewidth]{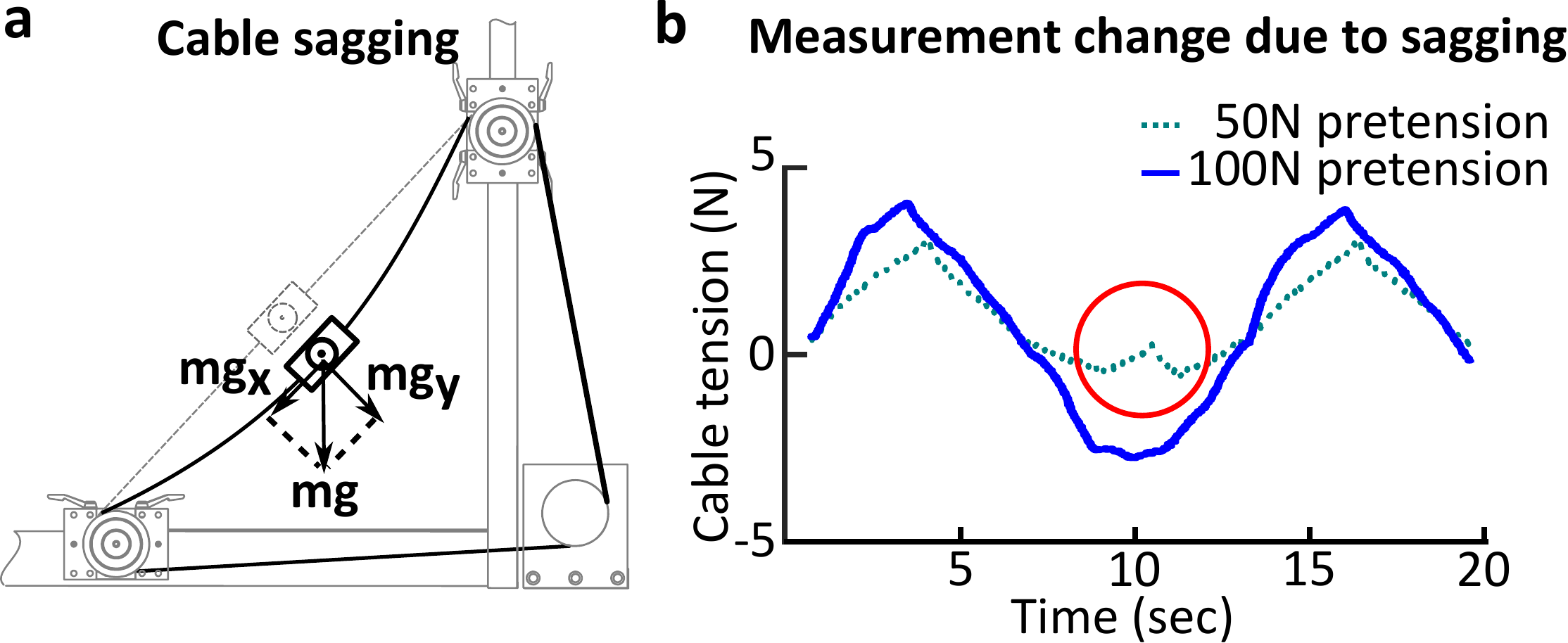}
\caption[Cable sagging]{\small Cable sagging. The weight of the cable system (i.e. cable, harness, load-cells and turnbuckles) deforms the cable as illustrated in (a). (b) During a back and forth motion, the measured force will not change monotonically, which is marked by a red circle. Cable sagging can be minimised by increasing the cable tension.}
\label{fig:CableCharacteristics}
\end{figure}

\subsection{Cable's tension spatial and temporal dependency}
\label{ch:Cable's tension spatial and temporal dependency}
The mass of the load-cells, connecting the elements and the cable itself (0.5\,kg) will slightly bend the cable and create cable sagging as shown in Fig.\,\ref{fig:CableCharacteristics}a. As the cable is not perfectly straight, we observe that the tension does not change monotonically (as shown in Fig.\,\ref{fig:CableCharacteristics}b with cable tension measured from a single load-cell during a back and forth motion) and this discontinuity in force measurements is caused by a misalignment between the load-cells' axis and the cable's motion. Trigonometric calculations showed that a pretension of 200\,N limits the relative error between the measured and actual tensions below 2.5\%, thus also limiting the stiffness prediction error to 2.5\%. All further experiments were then performed with 200\,N pretension. 

NED's cable tension will slightly change over time due to the nature of the turnbuckles. During the validation tests, it was observed that the measured cable tension will drop by 1\,N every 33.6\,s (during a cyclic movement test of speed 750\,mm/s with a pretension value of 200\,N while holding an 18\,kg dummy leg, which will be described in Section~\ref{ch:Dummy leg mechanics}). This tension drop is negligible since most movements for neuromechanics evaluation require short perturbations with duration $<$1\,s (as will be developed in Section \ref{ch:Stiffness estimation}). The cable temporal dependency is described in \cite{Huang2019c}.

%%%%%%%%%%%%%%%
%figure
\begin{figure}[!b]
\centering
\includegraphics[width=0.99 \columnwidth]{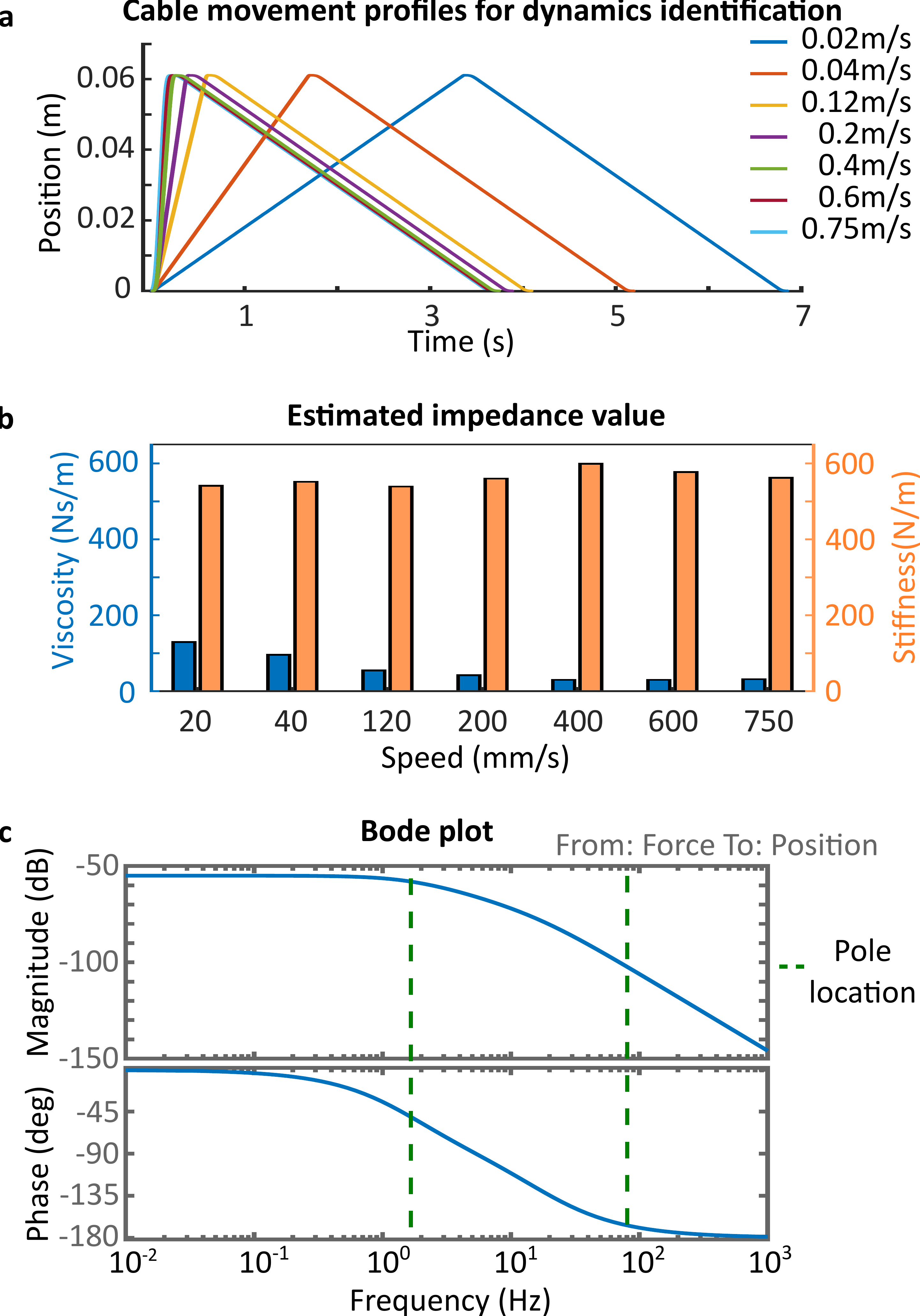}
\caption[Identification of NED as a linear second order system]{\small Identification results of NED as a linear second order system. Panel (a) depicts the perturbation profiles at different speed with estimated impedance values shown in panel (b). Panel (c) is the Bode plot of the system (with average values) that contains two poles at 1.6 and 17.2Hz (marked with dashed line). }
\label{fig:FitResults}
\end{figure}

\subsection{Cable system modelling}
\label{ch:Cable system modelling}
We performed system identification tests to demonstrate that the behaviour of the designed interface can be characterised by a second order linear dynamical system under different speed. Hence, the transfer function describing the cable's dynamics with cable tension, $\Delta F(s)$, as input, and cable displacement, $\Delta X(s)$, as output can be expressed as:
\begin{equation}
\label{eq:MechanicalTransferfunction}
\frac{\Delta X(s)}{\Delta F(s)} = \frac{1}{M_x s^2 + B_x s + K_x}
\end{equation}
with $M_x$ the mass of the moving components on the cable, $B_x$ and $K_x$ the cable viscosity and stiffness, respectively.

For system identification tests, we pre-programmed NED's controller to perform 10 saw-shape displacement patterns with $\pm$60\,mm amplitude and speeds of 20\,-\,750\,mm/s as shown in Fig.\,\ref{fig:FitResults}a. The force acting on the cable was measured with a load-cell and recorded at 1\,kHz. All ten trials at a given speed condition were used to estimate the transfer function~\eqref{eq:MechanicalTransferfunction} using a least square method {\it tfest} of Mathworks Matlab. Additionally, the mass of all moving mechanical components was known and was used as an initial estimate for mass component ($M_x$) of transfer function~\eqref{eq:MechanicalTransferfunction}.

The parameters identified at different speeds are shown in Fig.\,\ref{fig:FitResults}b, demonstrating that our interface is characterised by high stiffness and low viscosity (which reduces with the speed). This indicates that (despite the inherent cable compliance) NED is a rigid device that can be used to identify the lower limb mechanics. The performance of the fitting was confirmed by normalised root mean square error value (NRMSE) with values higher than 70\%. Due to both low variance in estimated impedance and relatively high NRMSE value, the mechanical characteristic of NED can be described by the average transfer function parameter's values, with the averaged Bode plot shown in Fig.\,\ref{fig:FitResults}c which has two poles, at 1.6\,Hz and 17.2\,Hz respectively. Identification of the cable's nonlinearities is described in \cite{Huang2019c}.

%%%%%%%%%%%%%%%%%%
\section{Validation}
\label{ch:Validating}
In this section, we demonstrate how NED can be used for lower limb neuromechanics characterisation. First, two experiments were conducted to identify the dynamic parameters of a dummy leg and a pair of springs separately, and then the experiment identification results were compared to known mechanic properties of the components. We then determined an optimal position perturbation for estimating the hip joint stiffness of healthy subjects.

%%%%%%%%%%%%%%%%%%
%figure
\begin{figure}[!b]
\centering
\includegraphics[width= \columnwidth]{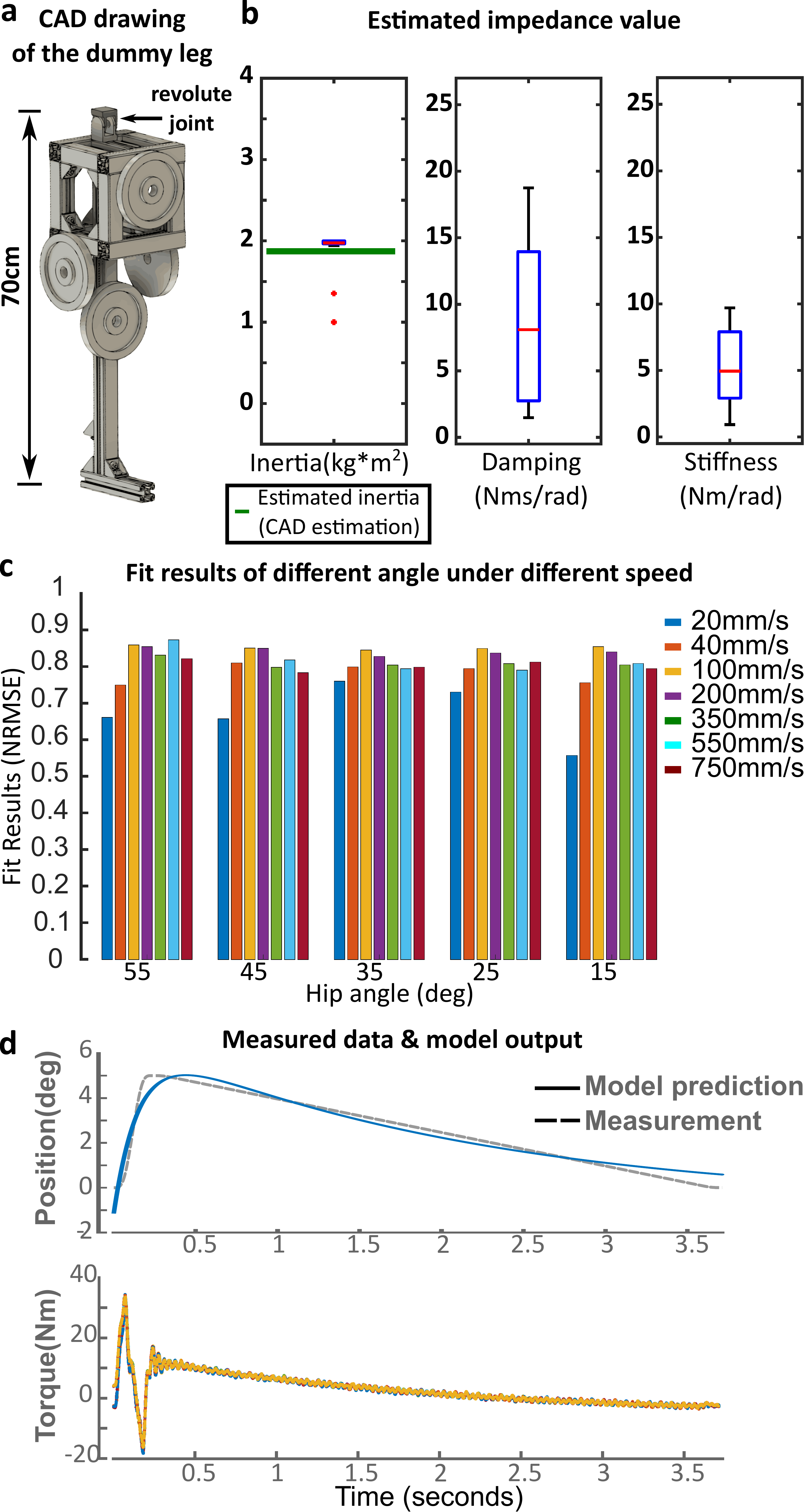}
\caption[Identification of the mechanics of a 18\,kg dummy leg at joint angles between 15$^\circ$-\,55$^\circ$ and speed 20\,-\,750\,mm/s, with 20 trials at each condition]{\small Identification of the mechanics of a 18\,kg dummy leg at joint angles between 15$^\circ$-\,55$^\circ$ and speed 20\,-\,750\,mm/s, with 20 trials at each condition. Panel (a) shows the CAD drawing of the dummy leg, Panel (b) the boxplot of the estimated inertia, damping and stiffness value of the dummy leg, with a green line indicating the estimated inertia from CAD (1.84\,kg\,m$^2$). Panel (c) shows the NRMSE value (which defines the estimation's quality) at five different hip angles and seven different speeds. Panel (d) shows the measurement results (dash line) and estimated model output (solid line) of the dummy leg identification.}
\label{fig:DummyLegFitResults}
\end{figure}

\subsection{Dummy leg mechanics}
\label{ch:Dummy leg mechanics}
To validate the functionality of the developed interface, experiments were carried to identify the mechanical dummy leg properties and compared with the values obtained from CAD calculations (Fig.\,\ref{fig:DummyLegFitResults}a). The design parameters of the leg were: mass 18\,kg (resembles the leg mass of a 90\,kg subject), length of 70\,cm and moment of inertia of 1.84\,kg\,m$^2$ with respect to the hip joint. This mechanical dummy leg was fixed in NED for neuromechanics experiments. During the experiment, the leg was rapidly displaced by an amplitude of 5$^\circ$ (6\,cm endpoint displacement) in both flexion and extension with speed ranging from 20\,-\,750\,mm/s, as shown in Fig.\,\ref{fig:DummyLegFitResults}d. This perturbation profile is similar to a ramp and hold motion. Using the dummy leg, we test the system performance and impedance estimation at dynamic conditions that could not be used with a human. Furthermore, the flexion/extension movements were tested at five different hip angles between 15\,$^{\circ}$ and 55\,$^{\circ}$ to study the influence of gravity on the identification results. In total, 20 repetitions were performed at each combined condition of speed [20\,-\,750]\,mm/s and hip angle [15$^\circ$-\,55$^\circ$].

As described in Section~\ref{ch:Cable system modelling}, the developed robot exhibits high stiffness and low viscosity which resembles a rigid device. Therefore, the cable dynamics in series with the leg could be neglected, and the recorded displacements and cable-leg interaction forces were used to estimate a linear second-order model of the mechanical dummy leg (with the least square method {\it tfest} of Mathworks Matlab): 
\begin{equation}
\label{eq:jointimpedance}
\Delta\tau = I\Delta\ddot{\theta} + B\Delta\dot{\theta} + K\Delta\theta\,, \,\,\, \Delta \tau = (F_1-F_2) L
\end{equation}
where $\Delta \tau$ is the change of interactive torque, $I$ the leg inertia, $B$ the viscous parameter of the joint, $K$ the hip joint stiffness, $F_{1}$ and $F_{2}$ are the two load-cells' signals. It is important to note that the change in interaction torque ($\Delta (F_1 - F_2)L$) was used here rather than change in one load-cell measurement ($\Delta F_1$), as the purpose of this section was to estimate dummy leg impedance through position displacement and resulting force changes. 

The estimated impedance values are shown in Fig.\,\ref{fig:DummyLegFitResults}b with the NRMSE value depicted in Fig.\,\ref{fig:DummyLegFitResults}c which suggests that the dummy leg's dynamic parameters were successfully identified for velocities larger than 40\,mm/s (with NRMSE $>$ 80\%). The inertia estimated in this dynamic identification is close to the value predicted from the CAD parameters (1.84\,kg\,m$^2$) while the viscosity and stiffness values are both low. These results indicate that NED can be used to identify the hip mechanical impedance.

%%%%%%%%%%%%%%%%%%
%figure
\begin{figure}[]
\centering
\includegraphics[width= 0.78\columnwidth]{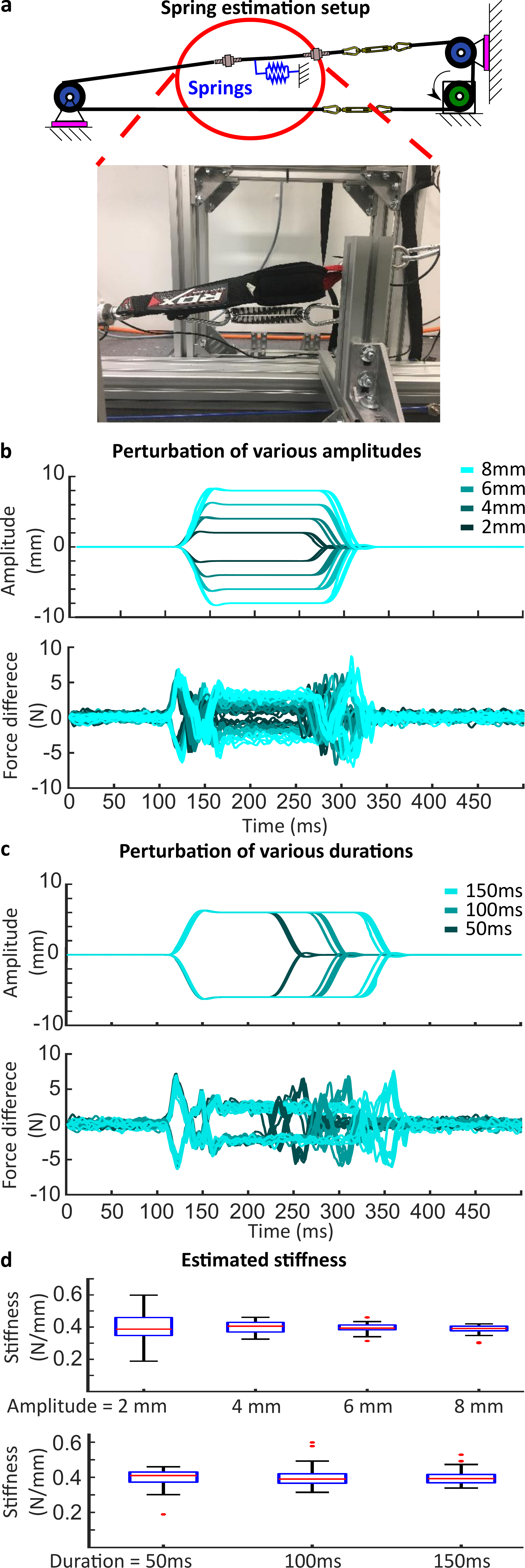}
\caption[Stiffness identification of a spring with known elasticity using NED]{\small Stiffness identification of a spring with known elasticity using NED. Panel (a) shows where the spring is attached. Panel (b) and (c) show the perturbations of different amplitude and duration together with the resulting force difference. Panel (d) shows the values obtained with these perturbations, which are similar to the spring constant with less deviation using a larger perturbation.}
\label{fig:SpringEstimation}
\end{figure}

\subsection{Stiffness estimation}
\label{ch:Stiffness estimation}
To evaluate whether NED can be used to identify stiffness, we used two parallel springs attached to the cable as shown in Fig.\,\ref{fig:SpringEstimation}a. The stiffness of these two springs was then identified using a position displacement with smooth ramps up and down \cite{Burdet2000}. In this method, stiffness $K$ can be computed from: 
\begin{equation}
\label{eq:stiffness}
\Delta \tau = K \Delta\theta~
\Leftrightarrow \Delta F = K \Delta X
\end{equation}
on the constant position plateau where inertia and viscosity have little influence, as $\Delta\dot{\theta}=\Delta\ddot{\theta}=0$, see Fig.\,\ref{fig:SpringEstimation}b. As indicated in the equation, small joint rotation $\Delta \theta$ can be approximated by linear displacement $\Delta X$, and torque difference $\Delta \tau$ can be approximated force difference $\Delta F$. Twenty tests were carried out for each perturbation with amplitudes of 2\,-\,8\,mm and durations 50\,-\,150\,ms. Stiffness was then evaluated from Equ.(\ref{eq:stiffness}) with mean displacement $\Delta\theta$ and mean measured torque $\Delta \tau $ during the plateau region. The results of Fig.\,\ref{fig:SpringEstimation}d demonstrate that this method can identify stiffness accurately, with estimations improved with a larger perturbation amplitude and no observable difference in different duration or perturbation direction. Since the load-cells' measurement suffers from noise with standard deviation of 0.29\,N and maximum 1\,N, a large amplitude with stronger spring force will increase the signal to noise ratio and improve the estimation. In the meantime, the angle measurement of the motor shaft has a resolution of 0.019$^\circ$ corresponding to a 0.35\,mm cable displacement and angle of 0.46$^\circ$ for a 90\,cm long leg. This implies that a small perturbation amplitude (e.g. 2\,mm amplitude) will suffer from measurement errors. 

%%%%%%%%%%%%%%%%%%
\subsection{Optimal position perturbation to identify stiffness}
\label{ch:Optimal position perturbation to identify stiffness}
A pilot study with one healthy subject (female, age: 21\,y, weight: 54\,kg, height: 172\,cm) was carried out to evaluate the feasibility of using NED to identify hip joint stiffness with the technique described in the previous section. The experimental protocol was approved by the Imperial College Research Ethics Committee. The subject was informed on the device and experiment, and signed an informed consent form prior to the experiment. The participant's weight and leg length (from the anterior superior iliac spine to the lateral malleolus) were measured to estimate the leg inertia. A lockable knee brace was used to fix the knee joint at 0$^\circ$ angle.

%figure
\begin{figure}[]
\centering
\includegraphics[width=0.9 \columnwidth]{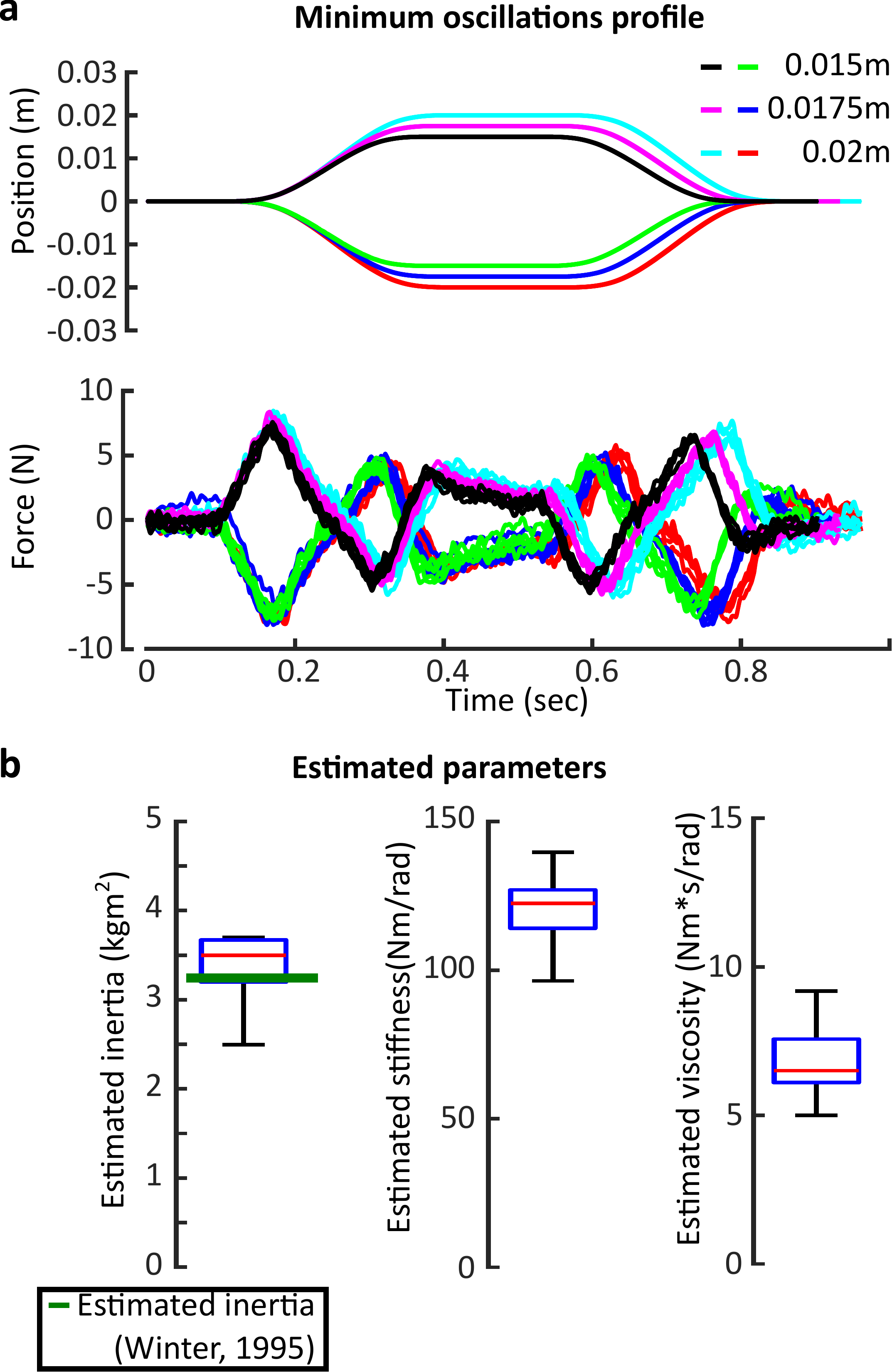}
\caption[Estimation result of a pilot study with one subject]{\small Estimation result of a pilot study with one subject. Panel (a) is an example of the profile with the position at the top and the force feedback at the bottom. Panel (b) shows the estimated joint impedance. The estimated inertia is close to the estimation from the anatomical model (Winter, 1995) shown is a green line.}
\label{fig:MinimumOscillation}
\end{figure}
The subject was half seated on the chair with one leg suspended, was asked to support his body weight on the handle and relax the lower limb. A harness was attached to the ankle of the tested leg which was connected to the cable and the motor (Fig.\,\ref{fig:NEDSketch}). The participant was given an emergency stop and could stop the experiment whenever needed. The laser safety system was initiated and adjusted to define the range of motion of the tested leg at 15$^\circ$. The motor performed a slow motion to move the participant's leg to define the comfortable range of motion for additional system adjustments. A single pulse perturbation was also given to provide the participant with an experience of a perturbation and adjust the system's safety.

As described in Section\,\ref{ch:Control system}, the motor controller implements the fastest displacement given the safety limits in speed, acceleration and jerk. The position perturbation with a constant displacement plateau was thus determined by these limits as well as by the plateau duration and displacement amplitude. Our goal was to use a perturbation of minimum amplitude and duration, which would disturb the subject minimally and thus avoid any voluntary reaction. However, this comes in trade-off with the perturbation amplitude, which should be large enough to maximise the signal to noise ratio (SNR) of the stiffness estimation. On the other hand, very large accelerations to yield a fast perturbation could cause cable oscillations which would disturb the subject and affect the stiffness estimation quality. To achieve fast displacement with minimal cable oscillations and stable force measurements, we increased the perturbation amplitude iteratively (starting from 6cm) and reduced the controller's dynamic limits (speed, acceleration and jerk) while recording the magnitudes of cable oscillation. In total, 50 combinations of the speed, acceleration and jerk limits were tested for position perturbation command with the hip flexed at 15$^\circ$.

The resulting optimal perturbation is shown in Fig.\,\ref{fig:MinimumOscillation}a. The amplitude of the perturbation is largely above the position resolution (0.35\,mm cable displacement) and from the test with the spring of Fig.\,\ref{fig:SpringEstimation}, should have a high signal to noise ratio. Considering the large motor variability in human movements, we selected a larger displacement than required to maintain a high signal to noise ratio. The collected kinematics and interaction forces were again least square fitted to estimate the hip joint impedance using Equ.(\ref{eq:jointimpedance}). As shown in Fig.\,\ref{fig:MinimumOscillation}a, the optimised perturbation resulted in consistent and reproducible motions with negligible force oscillations.

Fig.\,\ref{fig:MinimumOscillation}b shows the estimated joint impedance ($I,B,K$) of three different perturbation amplitudes \{15,\,17.5,\,20\}\,mm with a 150\,ms long plateau. The green line shown in Fig.\,\ref{fig:MinimumOscillation}b is the inertia calculated from the anatomical model \cite{Winter2009} using the subject's weight and leg length. We can see that the variance of estimation is small (11\% for inertia, 10\% for stiffness and 19\% for viscosity). The inertia estimate is close to the anatomical model, and the values of viscoelasticity are in the same order of magnitude as reported in \cite{Koopman2016}. 

%%%%%%%%%%%%%%%%%%
\section{Discussion and Conclusion}
Investigating the lower-limb neuromechanics is important to understanding the control of standing and walking in healthy and neurologically affected individuals, as well as to efficiently control robotic devices for assistance, rehabilitation and augmentation. However, so far few studies could use a single device to investigate lower-limb neuromechanics of different joints and specific to the hip. Importantly, hip joint viscoelasticity investigation was only performed in multi-joint torque perturbation \cite{Koopman2016}, which is usually limited in accuracy, and never with precise single joint position displacement. In this context, we have developed and validated a novel robotic interface named NED (Neuromechanics Evaluation Device) to investigate the lower-limb neuromechanics.

NED can apply a large range of dynamic interactions to a subject's leg at static posture or during movement. This enables the neuromechanics identification of hip and knee joints in flexion/extension. Importantly, NED allows the experimenter to estimate a subject lower limb neuromechanics in a natural upright posture under a controlled environment, which also makes the device well suited for carrying out investigations on patients' neuromechanics. The device can be quickly adapted to a subject's specific anatomy and to carry out various measurements. The use of a closed mechanical cable loop with a powerful actuator fixed outside the rigid supporting structure enables us to implement highly dynamic environments with little vibrations.

In this paper, NED's mechanics was characterised, and its performance to estimate a lower limb neuromechanics was demonstrated through the identification of a dummy leg and a spring's mechanical impedance. As a result of a powerful actuator and stiff mechanical frame of NED, it was possible to achieve accurate and repeatable position perturbation which enabled more efficient dynamics identification of individual leg joint compared to the mechanisms with rigid links~\cite{Hahn2011, Fujii2007}. The good match of the identified and measured parameters as well as the range of protocols that can be implemented on NED makes it an effective tool to identify the hip, knee and ankle joint biomechanics. The techniques developed in this paper could be used to systematically investigate hip joint viscoelasticity as described in \cite{Huang2020a}.

%%%%%%%%%%%%%%%
\bibliographystyle{IEEEtran}
\bibliography{bibliography}

\begin{thebibliography}{10}
\providecommand{\url}[1]{#1}
\csname url@samestyle\endcsname
\providecommand{\newblock}{\relax}
\providecommand{\bibinfo}[2]{#2}
\providecommand{\BIBentrySTDinterwordspacing}{\spaceskip=0pt\relax}
\providecommand{\BIBentryALTinterwordstretchfactor}{4}
\providecommand{\BIBentryALTinterwordspacing}{\spaceskip=\fontdimen2\font plus
\BIBentryALTinterwordstretchfactor\fontdimen3\font minus
  \fontdimen4\font\relax}
\providecommand{\BIBforeignlanguage}[2]{{%
\expandafter\ifx\csname l@#1\endcsname\relax
\typeout{** WARNING: IEEEtran.bst: No hyphenation pattern has been}%
\typeout{** loaded for the language `#1'. Using the pattern for}%
\typeout{** the default language instead.}%
\else
\language=\csname l@#1\endcsname
\fi
#2}}
\providecommand{\BIBdecl}{\relax}
\BIBdecl

\bibitem{Sinclair2006}
P.~J. Sinclair, G.~M. Davis, and R.~M. Smith, ``{Musculo-skeletal modelling of
  NMES-evoked knee extension in spinal cord injury},'' \emph{Journal of
  Biomechanics}, vol.~39, no.~3, pp. 483--92, 2006.

\bibitem{Hahn2011}
D.~Hahn, M.~Olvermann, J.~Richtberg, W.~Seiberl, and A.~Schwirtz, ``{Knee and
  ankle joint torque-angle relationships of multi-joint leg extension},''
  \emph{Journal of Biomechanics}, vol.~44, no.~11, pp. 2059--65, 2011.

\bibitem{Amankwah2004}
K.~Amankwah, R.~J. Triolo, and R.~F. Kirsch, ``{Effects of spinal cord injury
  on lower-limb passive joint moments revealed through a nonlinear viscoelastic
  model.}'' \emph{Journal of Rehabilitation Research and Development}, vol.~41,
  no.~1, pp. 15--32, 2004.

\bibitem{Dirnberger2012b}
J.~Dirnberger, H.-P. Wiesinger, A.~K{\"{o}}sters, and E.~M{\"{u}}ller,
  ``{Reproducibility for isometric and isokinetic maximum knee flexion and
  extension measurements using the IsoMed 2000-dynamometer},''
  \emph{Isokinetics and Exercise Science}, vol.~20, no.~3, pp. 149--53, 2012.

\bibitem{Alvares2015}
J.~B. {de Araujo Ribeiro Alvares}, R.~Rodrigues, R.~{de Azevedo Franke},
  B.~G.~C. da~Silva, R.~S. Pinto, M.~A. Vaz, and B.~M. Baroni, ``{Inter-machine
  reliability of the Biodex and Cybex isokinetic dynamometers for knee
  flexor/extensor isometric, concentric and eccentric tests},'' \emph{Physical
  Therapy in Sport}, vol.~16, no.~1, pp. 59--65, 2015.

\bibitem{Valovich-mcLeod2004}
T.~C. Valovich-mcLeod, S.~J. Shultz, B.~M. Gansneder, D.~H. Perrin, and J.~M.
  Drouin, ``{Reliability and validity of the Biodex system 3 pro isokinetic
  dynamometer velocity, torque and position measurements},'' \emph{European
  Journal of Applied Physiology}, vol.~91, no.~1, pp. 22--9, 2004.

\bibitem{Lunenburger2005}
L.~L{\"{u}}nenburger, G.~Colombo, R.~Riener, and V.~Dietz, ``{Clinical
  assessments performed during robotic rehabilitation by the gait training
  robot Lokomat},'' in \emph{Proceedings of the IEEE International Conference
  on Rehabilitation Robotics}, 2005, pp. 345--8.

\bibitem{Veneman2007}
J.~F. Veneman, R.~Kruidhof, E.~E. Hekman, R.~Ekkelenkamp, E.~H. {Van
  Asseldonk}, and H.~{Van Der Kooij}, ``{Design and evaluation of the LOPES
  exoskeleton robot for interactive gait rehabilitation},'' \emph{IEEE
  Transactions on Neural Systems and Rehabilitation Engineering}, vol.~15,
  no.~3, pp. 379--86, sep 2007.

\bibitem{Lorentzen2010}
J.~Lorentzen, M.~J. Grey, C.~Crone, D.~Mazevet, F.~Biering-S{\o}rensen, and
  J.~B. Nielsen, ``{Distinguishing active from passive components of ankle
  plantar flexor stiffness in stroke, spinal cord injury and multiple
  sclerosis},'' \emph{Clinical Neurophysiology}, vol. 121, no.~11, pp.
  1939--51, 2010.

\bibitem{Perell1996}
K.~Perell, A.~Scremin, O.~Scremin, and C.~Kunkel, ``{Quantifying muscle tone in
  spinal cord injury patients using isokinetic dynamometric techniques.}''
  \emph{Paraplegia}, vol.~34, no.~1, pp. 46--53, 1996.

\bibitem{Akman1999}
M.~N. Akman, R.~Bengi, M.~Karatas, S.~Kilin{\c{c}}, S.~S{\"{o}}zay, and
  R.~Ozker, ``{Assessment of spasticity using isokinetic dynamometry in
  patients with spinal cord injury.}'' \emph{Spinal Cord}, vol.~37, no.~9, pp.
  638--43, 1999.

\bibitem{DeGooijer-VanDeGroep2013}
K.~L. {De Gooijer-Van De Groep}, E.~{De Vlugt}, J.~H. {De Groot}, H.~C. {Van
  Der Heijden-Maessen}, D.~H. Wielheesen, R.~S. {Van Wijlen-Hempel}, J.~H.
  Arendzen, and C.~G. Meskers, ``{Differentiation between non-neural and neural
  contributors to ankle joint stiffness in cerebral palsy},'' \emph{Journal of
  NeuroEngineering and Rehabilitation}, vol.~10, no.~1, p.~81, 2013.

\bibitem{Sloot2015}
L.~H. Sloot, M.~M. van~der Krogt, K.~L. d. G.-v. de~Groep, S.~van Eesbeek,
  J.~de~Groot, A.~I. Buizer, C.~Meskers, J.~G. Becher, E.~de~Vlugt, and
  J.~Harlaar, ``{The validity and reliability of modelled neural and tissue
  properties of the ankle muscles in children with cerebral palsy},''
  \emph{Gait and Posture}, vol.~42, no.~1, pp. 7--15, 2015.

\bibitem{Claiborne2009}
T.~L. Claiborne, M.~K. Timmons, and D.~M. Pincivero, ``{Test - retest
  reliability of cardinal plane isokinetic hip torque and EMG},'' \emph{Journal
  of Electromyography and Kinesiology}, vol.~19, no.~5, pp. e345--52, 2009.

\bibitem{Jin2015}
X.~Jin, X.~Cui, and S.~K. Agrawal, ``{Design of a cable-driven active leg
  exoskeleton (C-ALEX) and gait training experiments with human subjects},'' in
  \emph{Proceedings of the IEEE International Conference on Robotics and
  Automation (ICRA)}, 2015, pp. 5578--5583.

\bibitem{Vouga2017}
T.~Vouga, R.~Baud, J.~Fasola, M.~Bouri, and H.~Bleuler, ``{TWIICE - A
  lightweight lower-limb exoskeleton for complete paraplegics},'' in \emph{IEEE
  International Conference on Rehabilitation Robotics}.\hskip 1em plus 0.5em
  minus 0.4em\relax IEEE Computer Society, aug 2017, pp. 1639--1645.

\bibitem{Koopman2016}
B.~Koopman, E.~H.~F. van Asseldonk, and H.~van~der Kooij, ``{Estimation of
  human hip and knee multi-joint dynamics using the LOPES gait trainer},''
  \emph{IEEE Transactions on Robotics}, vol.~32, no.~4, pp. 920--32, 2016.

\bibitem{Mirbagheri2000}
M.~Mirbagheri, H.~Barbeau, and R.~Kearney, ``{Intrinsic and reflex
  contributions to human ankle stiffness: variation with activation level and
  position},'' \emph{Experimental Brain Research}, vol. 135, no.~4, pp.
  423--36, 2000.

\bibitem{Mirbagheri2001}
M.~Mirbagheri, H.~Barbeau, M.~Ladouceur, and R.~Kearney, ``{Intrinsic and
  reflex stiffness in normal and spastic, spinal cord injured subjects},''
  \emph{Experimental Brain Research}, vol. 141, no.~4, pp. 446--59, 2001.

\bibitem{Rouse2014}
E.~J. Rouse, L.~J. Hargrove, E.~J. Perreault, and T.~A. Kuiken, ``{Estimation
  of human ankle impedance during the stance phase of walking},'' \emph{IEEE
  Transactions on Neural Systems and Rehabilitation Engineering}, vol.~22,
  no.~4, pp. 870--8, 2014.

\bibitem{Chung2004}
S.~G. Chung, E.~{Van Rey}, Z.~Bai, E.~J. Roth, and L.~Q. Zhang, ``{Biomechanic
  changes in passive properties of hemiplegic ankles with spastic
  hypertonia},'' \emph{Archives of Physical Medicine and Rehabilitation},
  vol.~85, no.~10, pp. 1638--46, 2004.

\bibitem{Shorter2019}
A.~L. Shorter and E.~J. Rouse, ``{Ankle Mechanical Impedance During the Stance
  Phase of Running},'' \emph{IEEE Transactions on Biomedical Engineering}, sep
  2019.

\bibitem{Bieryla2009}
K.~A. Bieryla, D.~E. Anderson, and M.~L. Madigan, ``{Estimations of relative
  effort during sit-to-stand increase when accounting for variations in maximum
  voluntary torque with joint angle and angular velocity},'' \emph{Journal of
  Electromyography and Kinesiology}, vol.~19, no.~1, pp. 139--44, 2009.

\bibitem{Dirnberger2012a}
J.~Dirnberger, A.~K{\"{o}}sters, and E.~M{\"{u}}ller, ``{Concentric and
  eccentric isokinetic knee extension: A reproducibility study using the IsoMed
  2000-dynamometer},'' \emph{Isokinetics and Exercise Science}, vol.~20, no.~1,
  pp. 31--5, 2012.

\bibitem{Meuleman2016}
J.~Meuleman, E.~{Van Asseldonk}, G.~{Van Oort}, H.~Rietman, and H.~{Van Der
  Kooij}, ``{LOPES II - Design and evaluation of an admittance controlled gait
  training robot with shadow-leg approach},'' \emph{IEEE Transactions on Neural
  Systems and Rehabilitation Engineering}, vol.~24, no.~3, pp. 352--63, 2016.

\bibitem{Andriacchi1980}
T.~P. Andriacchi, G.~B. Andersson, R.~W. Fermier, D.~Stern, and J.~O. Galante,
  ``{A study of lower-limb mechanics during stair-climbing.}'' \emph{The
  Journal of Bone and Joint Surgery. American volume}, vol.~62, no.~5, pp.
  749--57, 1980.

\bibitem{Fornusek2007}
C.~Fornusek, P.~J. Sinclair, and G.~M. Davis, ``{The force-velocity
  relationship of paralyzed quadriceps muscles during functional electrical
  stimulation cycling},'' \emph{Neuromodulation}, vol.~10, no.~1, pp. 68--75,
  2007.

\bibitem{Winter2009}
D.~A. Winter, \emph{{Biomechanics and motor control of human movement}}.\hskip
  1em plus 0.5em minus 0.4em\relax Wiley, 2009.

\bibitem{Aoyagi2007}
D.~Aoyagi, W.~Ichinose, S.~Harkema, D.~Reinkensmeyer, and J.~Bobrow, ``{A robot
  and control algorithm that can synchronously assist in naturalistic motion
  during body-weight-supported gait training following neurologic injury},''
  \emph{IEEE Transactions on Neural Systems and Rehabilitation Engineering},
  vol.~15, no.~3, pp. 387--400, sep 2007.

\bibitem{Fujii2007}
A.~Fujii, S.~Oda, S.~Komada, and J.~Hirai, ``{A muscle tension estimation
  method by using mechanical impedance of human knee joint during training by
  manipulator},'' in \emph{Mechatronics,IEEE International Conference on
  Mechatronics}.\hskip 1em plus 0.5em minus 0.4em\relax IEEE, 2007, pp. 1--6.

\bibitem{Townsend1988}
W.~T. Townsend, ``{The effect of transmission design on force-controlled
  manipulator performance},'' Ph.D. dissertation, Massachusetts Institute of
  Technology, USA, 1988.

\bibitem{Huang2019c}
H.-Y. Huang, ``{Development of the neuromechanics evaluation device (NED) for
  subject-specific lower limb modelling of spinal cord injury},'' Ph.D.
  dissertation, Imperial College of Science, Technology and Medicine, London,
  UK, 2019.

\bibitem{Burdet2000}
E.~Burdet, R.~Osu, D.~W. Franklin, T.~Yoshioka, T.~E. Milner, and M.~Kawato,
  ``{A method for measuring endpoint stiffness during multi-joint arm
  movements.}'' \emph{Journal of Biomechanics}, vol.~33, no.~12, pp. 1705--9,
  2000.

\bibitem{Huang2020a}
H.-Y. Huang, A.~Arami, I.~Farkhatdinov, D.~Formica, and E.~Burdet, ``{The
  influence of posture, applied force and perturbation direction on hip joint
  viscoelasticity},'' \emph{IEEE Transactions on Neural Systems and
  Rehabilitation Engineering}, 2020.

\end{thebibliography}

\end{document}